\documentclass[letterpaper, 10 pt, journal, twoside]{IEEEtran} 

\usepackage{cite} 
\usepackage{hyperref}
\usepackage[utf8]{inputenc} 
\usepackage[misc]{ifsym}
\usepackage{graphicx}
\usepackage{makecell}
\usepackage{tikz}
\graphicspath{ {./images/} }
\usepackage{multirow}

\IEEEoverridecommandlockouts                              

\usepackage{amsmath} 
\begin{document}

\title{\LARGE \bf
Real-Time Instrument Segmentation in Robotic Surgery using Auxiliary Supervised Deep Adversarial Learning
}

\author{Mobarakol Islam, Daniel A. Atputharuban, Ravikiran Ramesh, and Hongliang Ren

\thanks{Manuscript received: September 10, 2018; Revised December 20, 2018; Accepted February 10, 2019.This paper was recommended for publication by Editor Tamim Asfour upon evaluation of the Associate Editor and Reviewers’ comments. This work was supported by the Singapore Academic Research Fund under Grant {R-397-000-297-114}, and NMRC Bedside \& Bench under grant R-397-000-245-511. (Corresponding author: Hongliang Ren)}

\thanks{M. Islam is with NUS Graduate School for Integrative Sciences and Engineering (NGS) and Dept. of Biomedical Engineering, National University of Singapore, Singapore
(e-mail: mobarakol@u.nus.edu)
        }
\thanks{D. A. Atputharuban is with Department of Electronics and Telecommunications, University of Moratuwa, Srilanka (e-mail: adanojan1@gmail.com) 
        }
\thanks{R. Ramesh is with Instrumentation and Control Engineering, NIT Trichy, India (e-mail: raviramesh.kiran97@gmail.com)
        }
\thanks{H. Ren is with Dept. of Biomedical Engineering, National University of Singapore, Singapore (e-mail: hlren@ieee.org/ren@nus.edu.sg)
        }
\thanks{Digital Object Identifier (DOI): see top of this page.}
}
\markboth{IEEE Robotics and Automation Letters. Preprint Version. Accepted February, 2019}
{Islam \MakeLowercase{\textit{et al.}}: Real-Time Instrument Segmentation in Robotic Surgery using Auxiliary Supervised Deep Adversarial Learning} 

\maketitle

\begin{abstract}
Robot-assisted surgery is an emerging technology which has undergone rapid growth with the development of robotics and imaging systems. Innovations in vision, haptics and accurate movements of robot arms have enabled surgeons to perform precise minimally invasive surgeries. Real-time semantic segmentation of the robotic instruments and tissues is a crucial step in robot-assisted surgery. Accurate and efficient segmentation of the surgical scene not only aids in the identification and tracking of instruments but also provided contextual information about the different tissues and instruments being operated with. For this purpose, we have developed a light-weight cascaded convolutional neural network (CNN) to segment the surgical instruments from high-resolution videos obtained from a commercial robotic system. We propose a multi-resolution feature fusion module (MFF) to fuse the feature maps of different dimensions and channels from the auxiliary and main branch. We also introduce a novel way of combining auxiliary loss and adversarial loss to regularize the segmentation model. Auxiliary loss helps the model to learn low-resolution features, and adversarial loss improves the segmentation prediction by learning higher order structural information. The model also consists of a light-weight spatial pyramid pooling (SPP) unit to aggregate rich contextual information in the intermediate stage. We show that our model surpasses existing algorithms for pixel-wise segmentation of surgical instruments in both prediction accuracy and segmentation time of high-resolution videos. 
\end{abstract}

\section{Introduction}
\IEEEPARstart{R}{obot-assisted} minimally invasive surgery (RMIS) has revolutionized the practice of surgery by optimizing surgical procedures, improving dexterous manipulations and enhancing patient safety \cite{taseWuLiaoPOE2014}.  Recent developments in the field of robotics, vision and smaller instruments have impacts on minimally invasive intervention. The common extensively used surgical robotic system is the Da Vinci Xi robot \cite{ freschi2013technical, ngu2017vinci,LiZhengMMTCompareTSMTCMCTM,FrServo14TASE} enable remote control laparoscopic surgery with long kinematic chains. The Raven II \cite{hannaford2013raven} is a robust surgical system consists of spherical positioning mechanisms. Remarkable recent surgical tools with complex actuation systems utilized micro-machined super-elastic tool \cite{devreker2015fluidic} and concentric tubes \cite{dwyer2017continuum}. However, with the reduction in size and complex actuation mechanisms, control of the instruments and cognitive representation of the robot kinematics are forthwith remarkably challenging in a surgical scenario. In addition, there are factors that complicate the surgical environment such as shadows and specular reflections, partial occlusion, smoke, and body fluid as well as the dynamic nature of background tissues. Hence, real-time surgical instruments detection, tracking, and isolation \cite{SS15MechShapeReconWDFMBezier,tubedet11iros,rn06iwasnrc,SY16RalSoftFabNov} from tissue are the key focus in the field of RMIS.

\begin{figure*}
\centering
\includegraphics[width=0.8\textwidth]{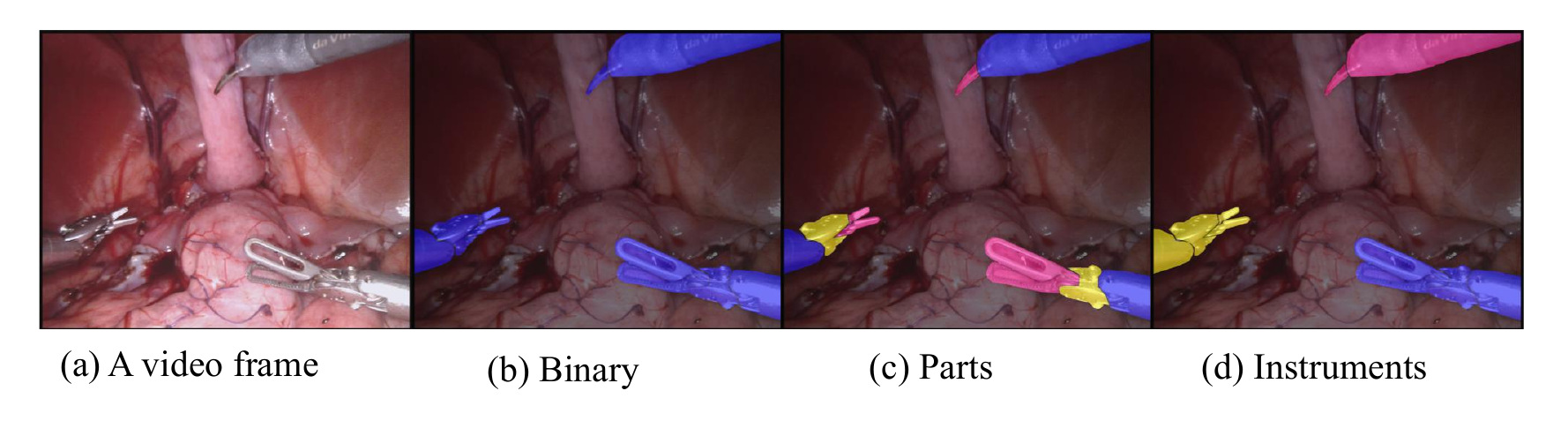}
\caption{Visualization of the robotic surgery image from the dataset that contains robotic instruments performing surgery on a tissue. The annotation of tools as binary (2 classes: Background and Instruments), parts (4 classes: Background, Shaft, Wrist, Claspers) and Instrument types (8 classes: Background, Bipolar Forceps, Prograsp Forceps, Large Needle Driver, Vessel Sealer, Grasping Retractor, Monopolar Curved Scissors, Other)}
\label{fig:Dataset_vis}
\end{figure*}

Previously, marker-based instruments tracking techniques apply in the robotic-assisted surgery \cite{SS15MechShapeReconWDFMBezier,tubedet11iros}. However, it increases the instrument's size and sterilization can be an issue in the MIS. Vision-based marker-free approaches for tracking are particularly desirable without increasing tools size on the existing setup. Prior methods utilize handcrafted features like color and texture features \cite{doignon2007segmentation, speidel2006tracking, zhou2014visual}, Haar wavelets \cite{sznitman2013unified}, HoG \cite{rieke2016real}, DFT shape matching \cite{su2018real} and some studies leverage classical machine learning models such as Random Forest \cite{bouget2015detecting}, Naive Bayesian \cite{speidel2006tracking} and Gaussian Mixture Model \cite{pezzementi2009articulated} to segment instrument's background. However, all these models are either solve a simple problem or not robust in intensity changes and typical motion of the instruments. Moreover, these models only apply for binary segmentation where it is necessary to detect parts and categories of the instruments to understand complex surgical scenario (see Fig. \ref{fig:Dataset_vis}).

Recently, deep learning has been excelled in the performance of the classification, detection and tracking problems. Semantic segmentation and tracking involving convolutional neural networks (CNN) have successfully been used in the field of medicine, for example, brain tumor segmentation \cite{bakas2018identifying, islam2017multi}, stroke lesion segmentation \cite{winzeck2018isles}, brain lesion segmentation \cite{kamnitsas2017efficient}, vessel tracking \cite{wu2016deep}, and tumor contouring \cite{terunuma2018novel}. 

\subsection{Related Work}
There are several successful deep learning approaches to localize and detect the pose and movement of instruments. To find the use of the real-time application, there are also few models focusing on prediction speed as well as accuracy.
Mostly, two type of studies for instruments tracking using CNN. First, tracking-by-detection using bounding box \cite{zhao2017tracking, chen2017surgical} and pose estimation \cite{rieke2016real}. However, bounding detection is not precise enough and seldom predicted locations are along instrument's body instead tip. Second, tracking-by-segmentation where instruments can be annotated into binary, parts and categories. ToolNet \cite{garcia2017toolnet}, a holistically nested real-time instrument segmentation approach of a robotic surgical tool. The work only focuses on binary segmentation with the observation of real-time prediction. 
Deep residual learning and dilated convolution are integrating to segment multi-class segmentation (instrument parts) and improve the binary segmentation  \cite{pakhomov2017deep}. Subsequently, Shvets et al. \cite{shvets2018automatic} segment the instruments into binary, parts and categories (the type of instruments) and further observe the prediction time for online application. The study uses the Jaccard index-based loss function to train LinkNet \cite{chaurasia2017linknet} and obtains better accuracy compared with other segmentation models. Laina et al.  \cite{laina2017concurrent} propose simultaneous segmentation and localization for tracking of surgical instruments. A pre-trained fully convolutional network (FCN) and affine transformation are used for non-rigid surgical tools tracking \cite{garcia2016real}. Another study \cite{al2018monitoring} checks the usage of the surgical tools by a joint model of CNN and recurrent neural network (RNN). Most of the approaches are attempting to track the instruments by emphasizing detection using convolutional networks which need tremendous computation. However, tracking instruments during surgery is an online task and it is crucial to supporting faster prediction speed for seamless surgery.

Online tasks such as instrument tracking during surgery are required an optimized model with good accuracy and prediction speed. There are very few works emphasize on fast semantic segmentation system with decent prediction performance from high-resolution video frames. ICNet \cite{zhao2018icnet} introduces cascade feature fusion (CFF) and auxiliary loss for real-time semantic segmentation. It leverages multiple branches with pyramid pooling and appends softmax cross-entropy loss in each branch. An encoder-decoder approach, LinkNet \cite{chaurasia2017linknet}, utilizes the model parameters efficiently and shows accurate instance level prediction without compromising processing time. Some other approaches such as ENet \cite{paszke2016enet}, SqueezeNet \cite{iandola2016squeezenet} trade-off accuracy and processing time by reducing filter size and input channels. Recently, adversarial learning models have been shown state of the art performance in the image synthesizing \cite{shin2018medical}, segmentation \cite{luc2016semantic} and tracking \cite{zhao2018adversarial}. Adversarial training optimizes objective function by adding adversarial term with conventional cross-entropy loss. It can enforce the higher-order consistency of the feature maps without changing model complexity.

\subsection{Contributions}
In this paper, we propose a light-weights CNN model with adversarial learning scheme for real-time surgical instruments segmentation from high-resolution videos. We have designed a multi-resolution feature fusion (MFF) module to aggregate the multi-resolution and multi-channel feature maps from auxiliary and master branches. We have also proposed a model regularization technique combining auxiliary and adversarial loss where auxiliary loss learns the low-resolution features and adversarial loss refines the higher order inconsistency of the feature maps. The proposed model further consists of convolution and deconvolution blocks, residual block, class block, decoder, and spatial pyramid pooling unit. To train in adversarial manners, we adopt an FCN followed by up-sampling layers as a discriminator \cite{luc2016semantic}. To enable real-time instruments tracking, we have tuned the model parameters and a trade-off between speed and accuracy to find out the optimized architecture. Our model has surpassed the performance of previous work on the MICCAI robotic instrument segmentation challenge 2017 \cite{allan20192017} in each category of segmentation such as binary, parts, and instruments. 

\section{Proposed Method}
Our proposed model consists of multiple branches over which contextual information from different resolutions of input images are fused to generate high-resolution semantic feature maps. We propose a Multi-resolution Feature Fusion (MFF) block to aggregate multi-scale features from a different branch. We also adopt spatial pyramid pooling where rich contextual features are reconstructed at different grid scales from bottom-up. Fig. \ref{fig:ours_segmentor} shows our proposed segmentation network of auxiliary (top) and main (bottom) branch and arrangement of different units such as Conv-Block, Residual-Block, MFF, Decoder, and Up-sampling. We refine predicted feature maps of our segmentation network by using a discriminator network in an adversarial learning manner, as illustrated in Fig. \ref{fig:dis}.
\subsection{Multi-resolution Feature Fusion (MFF)}

\begin{figure}[!ht]
\centering
\includegraphics[width=0.48\textwidth]{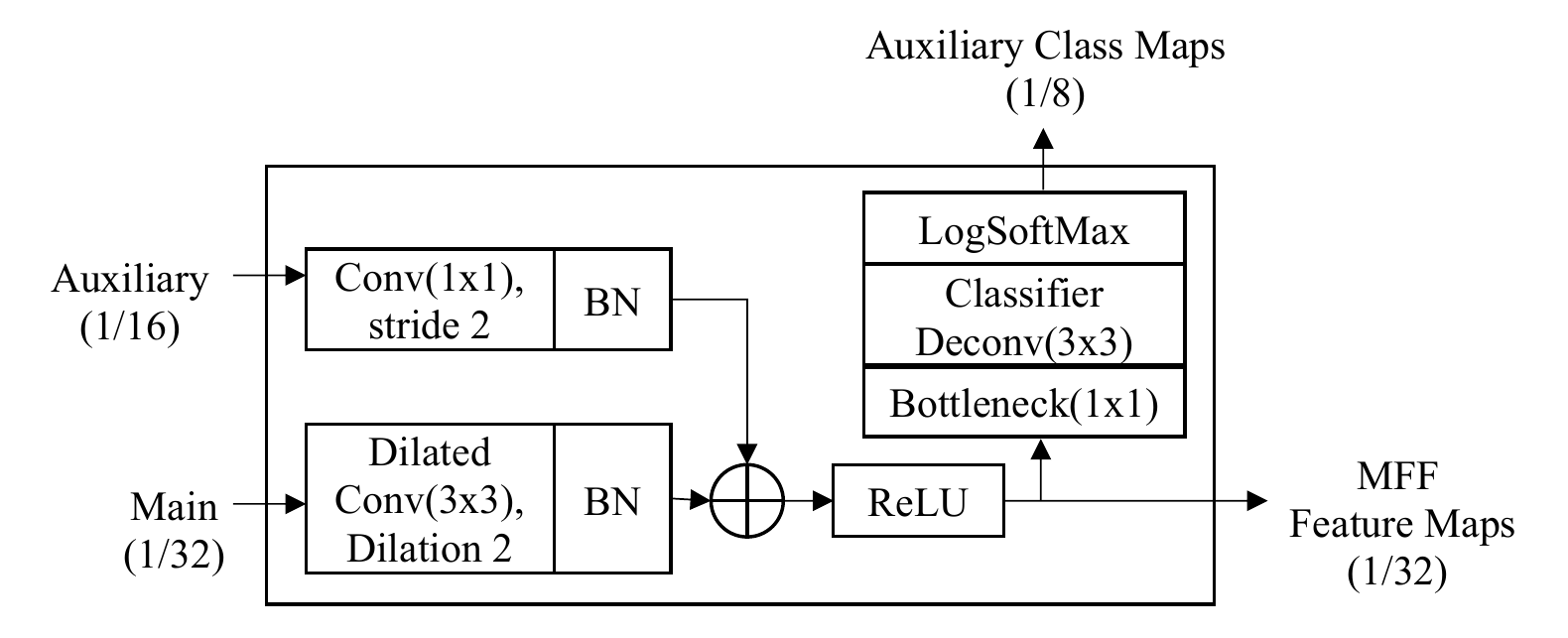}
\caption{Our Proposed Multi-resolution Feature Fusion (MFF) Module. Feature maps of Auxiliary branch (1/16) are downsampled and fused with main branch (1/32) and produced MFF feature maps and auxiliary class maps.}
\label{fig:mff}
\end{figure}

\begin{figure*}[!htbp]
\centering
\includegraphics[width=1\textwidth]{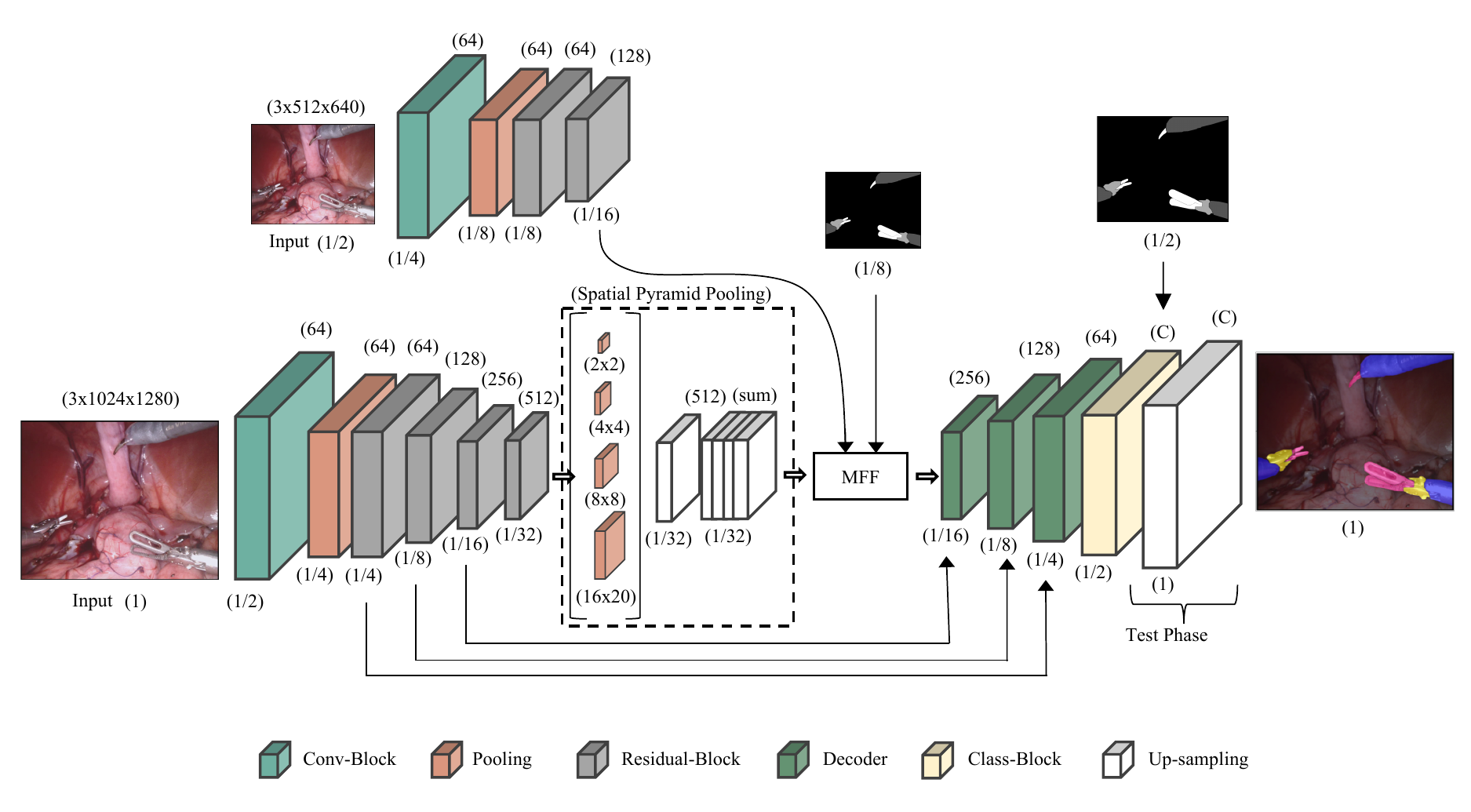}
\caption{Our proposed segmentation network. It has 2 branches with the different resolution of inputs. The feature maps of both branches are fused by proposed Multi-resolution feature fusion (MFF) module. In training time, the main loss calculated on (1/2) of the original resolution. Feature maps have been upsampled to 2x to fit with original dimension in the testing phase.}
\label{fig:ours_segmentor}
\end{figure*}

To combine the feature maps of different dimensions from main and auxiliary branches, we design multi-resolution feature fusion (MFF) module, as illustrated in Fig. \ref{fig:mff}. MFF can also produce the auxiliary class maps to calculate auxiliary loss. We adopt the idea of CFF from ICNet \cite{zhao2018icnet}. However, we replace the interpolation layer (upsample) with convolution layer (stride 2) to downsample the maps and added bottleneck layer to reduce channel without increasing complexity. We deal with various dimensions and channels of feature maps from multiple branches with MFF where CFF only works on different dimensions.

There are two inputs of scale 1/16 (auxiliary) and 1/32 (main) to MFF module where it downsamples auxiliary inputs and fuses with feature maps of the main branch. Auxiliary class maps and fused feature maps are the two outputs of the module.    
 
\subsection{Network Architecture}
\begin{figure*}[!htbp]
\centering
\includegraphics[width=0.7\textwidth]{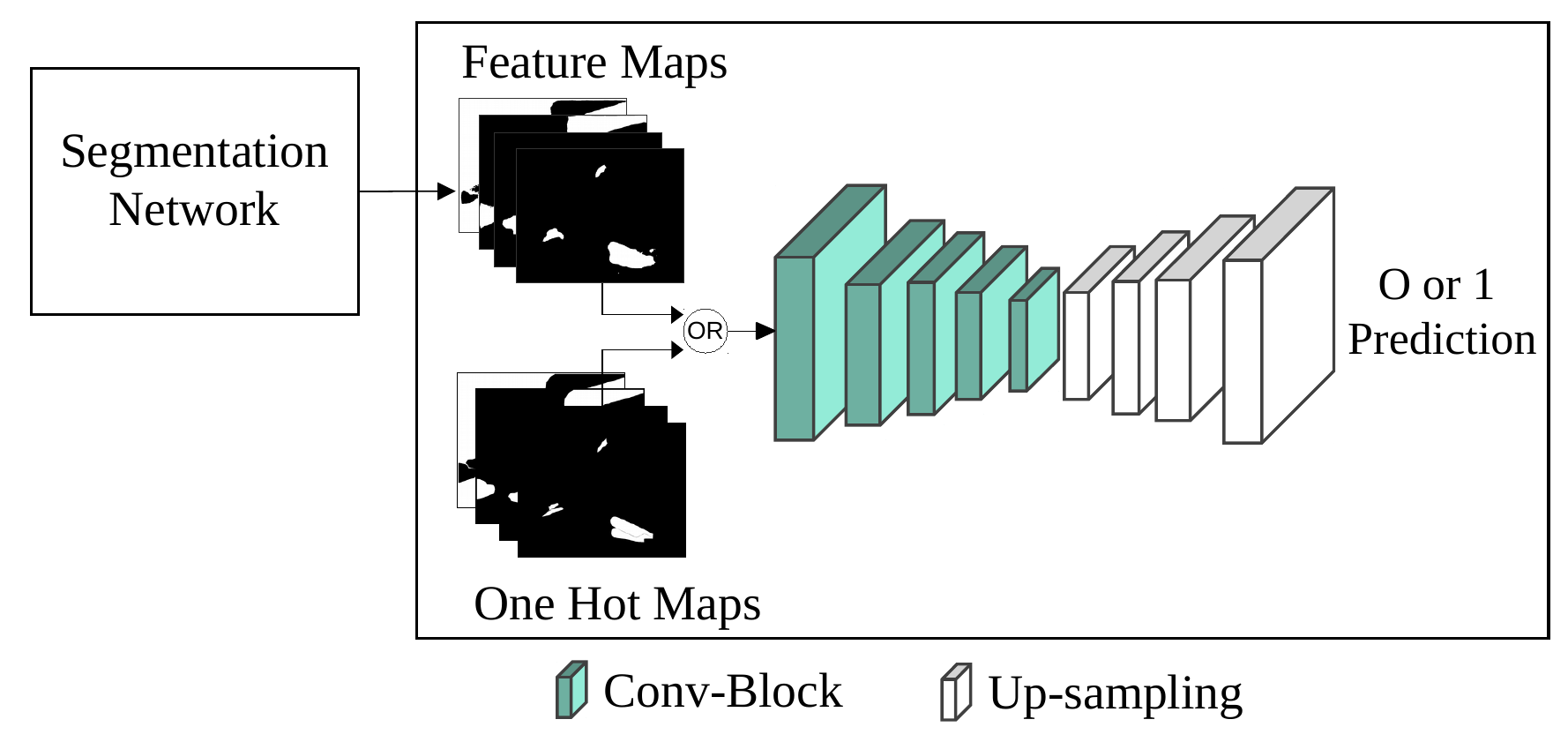}
\caption{Our proposed segmentation framework with adversarial learning scheme. Discriminator has 5 convolution layers followed by upsample layers.}
\label{fig:dis}
\end{figure*}

In Fig. \ref{fig:ours_segmentor}, the main branch consists of a Conv-Block followed by a max-pooling layer, 4 Residual-block, and a spatial pyramid pooling (SPP) unit. Conv-Block is the starting unit which forms with the layers of convolution, batch-normalization, and ReLU. It performs convolution on high-resolution input frames scale 1 such as 3x1024x1280 with a kernel size of 7 x 7 and stride of 2. There is a max-pooling immediately after Conv-Block to downsample the feature map into the half. Subsequently, there is 4 Residual-Blocks similar combination of layers as ResNet18 \cite{he2016deep} which is lighter and optimized with computation and accuracy. The quantity and scale of feature maps of each layer are depicted in the top and bottom respectively (Fig. \ref{fig:ours_segmentor}). A spatial pyramid pooling (SPP) \cite{zhao2017pyramid} unit utilizes to extract multi-scale semantic features from the output feature maps of the Residual-Blocks. To reduce feature length, we replace the concatenation operation of the pyramid pooling module with summation. The center of the segmentation architecture consists of MFF module which fuses the feature maps and produces auxiliary class maps. The latter part of the architecture has 3 decoder blocks and a class block similar to LinkNet \cite{chaurasia2017linknet}. Each decoder forms of Convolution (1x1)-Deconvolution (3x3, stride 2)-Convolution(1x1) followed by batch-norm and ReLU layers. There are also 3 layers inside the class block which connected as Deconvolution (3x3)-Convolution (3x3)-Deconvolution (2x2). To recover spatial information lost in downsampling, there is skip connection to each decoder from corresponding residual block.
The overall framework of our proposed model is depicted in Fig. \ref{fig:dis}. Generated feature maps from segmentation network and One-hot maps from ground truth are the input to the discriminator network. The network can differentiate the maps belongs to the segmentation network or ground truth and refine the high-level inconsistency. There are 5 Conv-Blocks and corresponding up-sampling (interpolation) layers in the discriminator network as \cite{hung2018adversarial}. The network can detect and correct the higher-order inconsistency of the predicted feature maps of the segmentation network.

\subsection{Loss Function}
The auxiliary loss at the intermediate stages helps to optimize the learning process and can be added with the main loss. It exploits the discrimination in low stages and provides more regularization in training. The segmentation loss ($L_{seg}$) function can be written as-
\begin{eqnarray}\label{eq:loss_aux}
L_{seg}= L_{main} + \lambda_{aux} L_{aux},
\end{eqnarray}
where $L_{main}$ and $L_{aux}$ are the softmax cross-entropy loss in main branch loss and auxiliary loss. We choose auxiliary weight factor $\lambda_{aux}$ = 0.4 as \cite{zhao2018icnet}. 

The later portion of our model is an adversarial loss which discriminates the feature maps of the segmentation network from label maps of the ground truth. Adversarial loss term penalized the mismatches in a higher ordered label such as a region labeled with certain class exceeds the threshold. Overall, training loss is the combination of the master and auxiliary branches loss with the adversarial loss.
\begin{eqnarray}\label{eq:loss_aux_adv}
L= L_{main} + \lambda_{aux} L_{aux}+ \lambda_{adv} L_{adv},
\end{eqnarray}
where $L_{adv}$ is the adversarial loss that is spatial cross entropy loss with respect to two classes (0 for feature maps of the segmentation network or 1 for label maps of the ground truth). We adopt the weight factor $\lambda_{adv} $ for the adversarial loss to be 0.01 as \cite{luc2016semantic}.

\section{Experiment}

\subsection{Dataset}The dataset used in this paper was provided by MICCAI 2017 as a part of the Endovis-Robotic instrument segmentation sub-challenge \cite{allan20192017}. The dataset consists of 225 frame sequences from 8 different surgeries acquired from the Da Vinci Xi surgical system (see the Fig.  \ref{fig:Dataset_vis}). Each sequence consists of surgery images from two RGB stereo channels recorded using the left and right camera respectively. For every image from the left camera, separate hand-labeled ground truth images are supplied for every individual instrument. The instruments can belong to either of the categories, namely rigid shafts, articulated wrists, clampers or miscellaneous instruments such as a laparoscopic instrument or drop-in ultrasound probe. Each image has a 1920 x 1080 resolution, which is reduced to 1280 x 1024 after cropping out the black canvas. For binary segmentation, we encode the value of 1 for every pixel that has an instrument and 0 for the background. For partwise segmentation, we encode every component of the instrument with values (0,1,2,3). For instrument segmentation, we encode every instrument category with an incremental numerical value starting at 1. 

We split the given training data into training and testing data. The image sequence from the first 6 surgeries consists of our training data, and contain a total of 1350 training images. The testing data consists of the image sequence from the remaining 2 surgeries and consists of a total of 450 images.

\subsection{Preprocessing}
The training dataset is augmented using simple augmentation (Flip Horizontal and Flip vertical) and the data set is normalized within each image channel by subtracting each channel's mean to get zero mean image. However, when the pre-trained model needs to be used for practical purposes, we can use additional augmentation techniques like Gaussian blur, Brightness change, and Image skew to simulate surgical conditions like fogging of the camera lens, changing of the brightness of input image and skewing of recording angle.

\subsection{Training}

We use 3 channel (RGB) endoscopy images and corresponding manually segmented images to train our model. The model is trained with Adam optimizer and the base learning rate of 0.001 for the segmentation network and 0.00015 for the discriminator. We adopt "poly" learning rate policy as \cite{chen2017rethinking}. Momentum is chosen to be 0.9 and weight decay term of 0.0005 used. We use Pytorch\cite{paszke2017automatic} deep learning platform to perform our experiments and the performance accuracy is calculated using the performance matrices given in Table \ref{table:with_adv}, \ref{table:binary_performance} and \ref{table:all_performance}. All the models train with 2 NVIDIA GTX 1080Ti GPU and inference time calculates on model prediction only excluding pre-processing and augmentation part. Batch size and number of GPU keep 1 in the inference phase so that we can have a fair comparison of speed.

\section{Results}

\begin{table}[!h]
\caption{Performance of our model with and without adversarial for binary Segmentation}
\begin{center}
\label{table:with_adv}
\begin{tabular}{|c|c|c|c|c|} 
\hline
 & Dice  & Hausdorff & Specificity & Sensitivity \\ \hline
With Adversarial & \textbf{0.916} & \textbf{11.11} & {0.989} & \textbf{0.928}\\ \hline
Without Adversarial  & 0.913 & 11.43 & \textbf{0.990} & 0.916\\ \hline

\end{tabular}
\end{center}
\end{table}

\begin{table}[!h]
\caption{Evaluation score for testing dataset of binary prediction. DR denotes as down-sampling rate for binary segmentation}
\begin{center}
\label{table:binary_performance}
\begin{tabular}{|c|c|c|c|c|c|c|} 
\hline
 & DR & Dice & Hausdorff &Specificity & Sensitivity \\ \hline
Ours & No & \textbf{0.916} & \textbf{11.110} & {0.989} & \textbf{0.928} \\ \hline
LinkNet \cite{shvets2018automatic} &No & {0.906} & 11.228 & {0.989} & {0.920}\\ \hline
ICNet \cite{zhao2018icnet} &No & 0.882 & {11.923}  & 0.986 & 0.892\\ \hline
UNet \cite{ronneberger2015u} &No &  0.878 & {12.112}  & 0.985 & 0.891\\ \hline
TernausNet \cite{iglovikov2018ternausnet} &No &  0.835 & {12.706}  & 0.983 & 0.830\\ \hline
PSPNet \cite{zhao2017pyramid}&2 & 0.831 &12.510  &\textbf{0.990} &0.788 \\ \hline

\end{tabular}
\end{center}
\end{table}

\begin{figure*}[!h]
\centering
\includegraphics[width=0.99\textwidth]{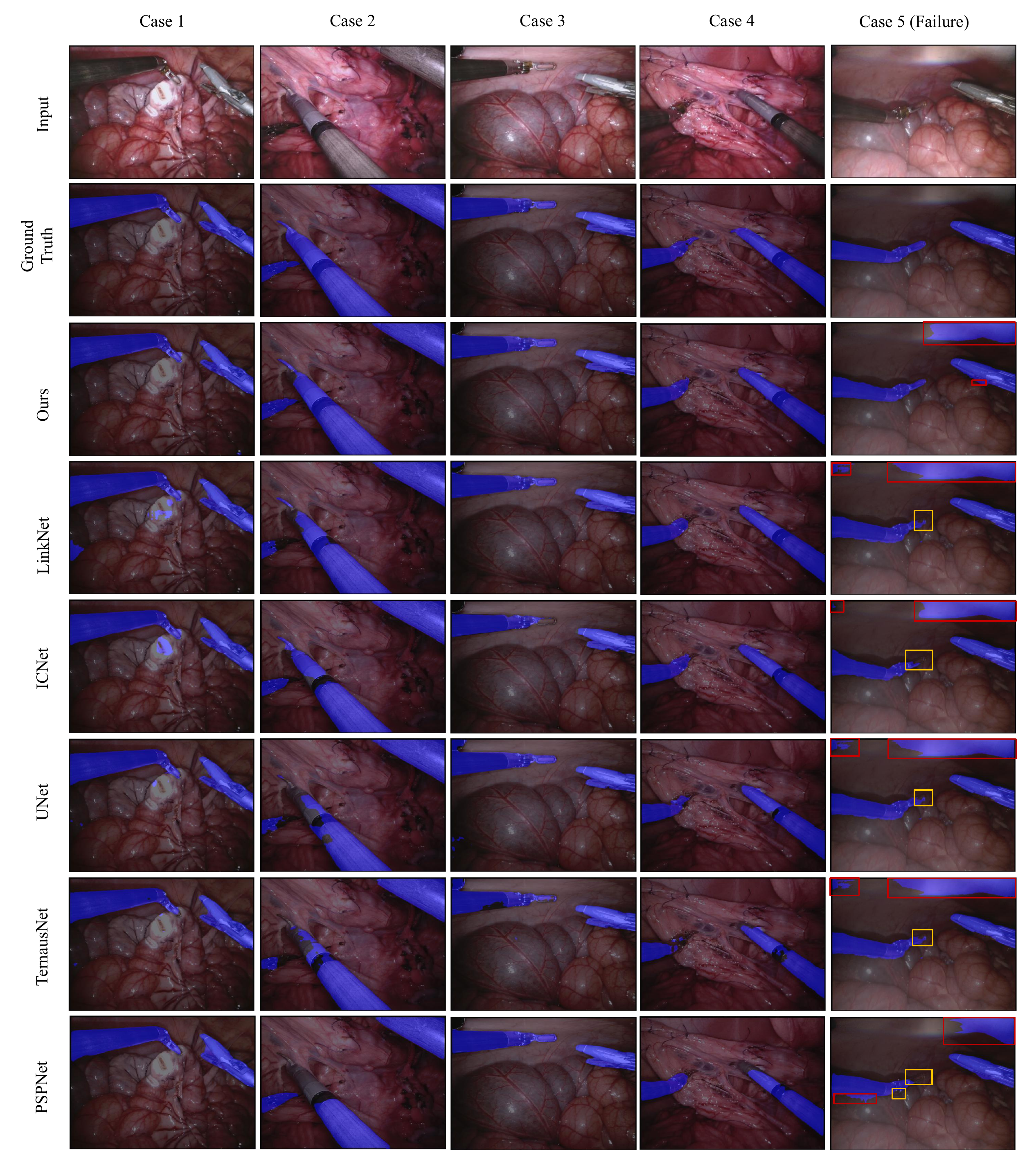}
\caption{Visualization of prediction results from different models. Cases from 1 to 4 are selected randomly. Predictions of our approach are comparable to the ground-truth whereas the predictions made by other models consist false positives and true negatives. Case 5 is one of the failure cases for our model where red and yellow boxes denote as the false positives and false negatives respectively.}
\label{fig:visual_comparison_binary}
\end{figure*}

\begin{figure*}[!h]
\centering
\includegraphics[width=0.99\textwidth]{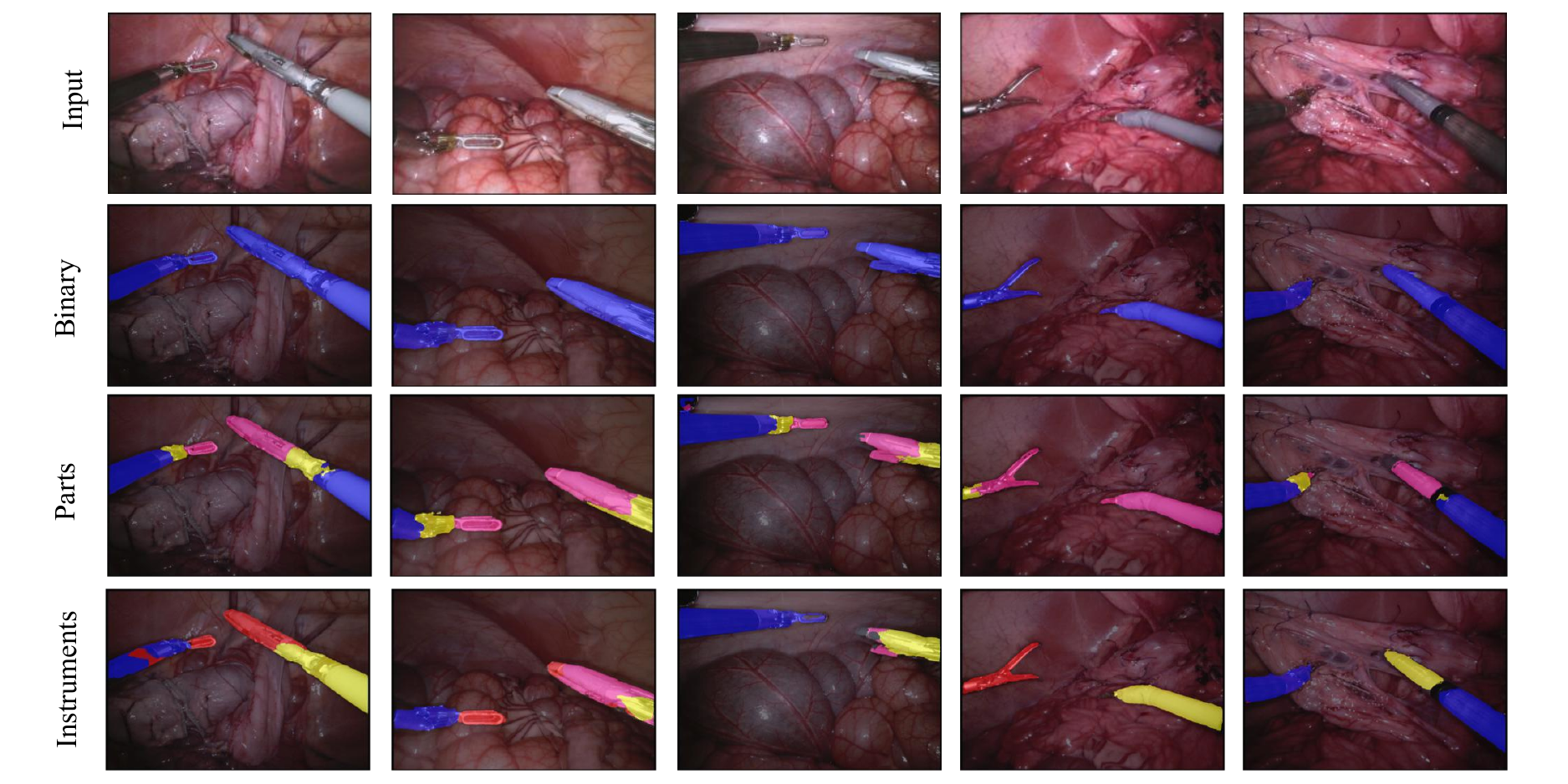}
\caption{Visualization of prediction results for the binary, parts and instruments wise segmentation. Proposed model shows high performance in binary and parts wise segmentation. There are many false positives predicted in instruments wise segmentation.}
\label{fig:visual_comparison}
\end{figure*}

The comparison of our model with existing architecture for binary, parts, and instruments wise segmentation is presented in Table \ref{table:with_adv}-\ref{table:all_performance} and Fig. 5 and 6. The visualization of binary segmentation of robotic instruments from background tissues is represented in Fig. \ref{fig:visual_comparison_binary}. Our model is close to the ground truth whereas there are false positive and true negatives in other architectures. In Table \ref{table:with_adv}, we have evaluated performance metrics for our segmentation architecture with and without adversarial learning. It’s evident that using adversarial learning results in better smoothens the class probabilities over the large region by enforcing spatial consistency. Table \ref{table:binary_performance} is the comparison of different models for the binary prediction on the testing data set. Our model achieves Dice and Hausdorff of 0.916 and 11.11 respectively which is almost a human level performance. This is the best results reported in literature up to now. In Table \ref{table:time}, we provide a comparison of time for prediction, training parameters and memory required. Though LinkNet \cite{shvets2018automatic} has shown the fastest model, but our model performs better in terms of accuracy(see the Table \ref{table:binary_performance} ). ICNet \cite{zhao2018icnet} requires minimum memory and number of parameters to train, but it also shows lower accuracy in parts and instruments segmentation (see Table \ref{table:all_performance}). In Table \ref{table:all_performance}, we present the results for binary, parts and instrument segmentation and we have visualized using Fig.\ref{fig:visual_comparison}. There are only 4 instruments (in total 7) used in the testing videos which could be the reason behind the lower segmentation accuracy of instrument categories. By investing dataset, we find that the missing instruments (Large Needle Driver and Prograsp Forceps) in the testing set are dominating the training sequences. LinkNet demonstrates competitive performance in all three segmentation types with the proposed model. Though UNet and ICNet also perform well in binary segmentation, they work poorly in parts and instruments segmentation. Overall, with the fps of 147.83 and best segmentation accuracy in binary, parts, and instruments segmentation our model has a clear edge over existing architectures.

\begin{table}[!h]
\caption{Average Time Consumed and required memory for Binary Prediction. Inference time measures on one NVIDIA GTX 1080Ti GPU and batch size 1}
\begin{center}
\label{table:time}
\begin{tabular}{|c|c|c|c|c|}
\hline
Model & \thead{Time\\(ms)} & fps & \thead{Memory \\(MB)}   & \thead{No. of Params\\(Millions)} \\ \hline
Ours & {5.75} & {173.78} & 81.8 & 14.91\\ \hline
LinkNet \cite{shvets2018automatic} & \textbf{4.07} & \textbf{245.88} & 46.2 & 11.79\\ \hline
ICNet \cite{zhao2018icnet} & 9.13 & 109.50 & \textbf{31.0} & \textbf{6.69}\\ \hline
UNet\cite{ronneberger2015u} & 4.46    & 224.21 & 31.4 & 7.84\\ \hline
TernausNet\cite{iglovikov2018ternausnet} & {4.20}    & {238.09} & 128.8 & 46.91\\ \hline
PSPNet\cite{zhao2017pyramid} & 16.25    &  61.55 & 272.8 & 68.05 \\ \hline
\end{tabular}
\end{center}
\end{table}

\begin{table}[!h]
\caption{Performance comparison for binary, instruments and parts segmentation with different models}
\begin{center}
\label{table:all_performance}
\begin{tabular}{|c|c|c|c|}
\hline
Model & Binary & Parts & Instruments \\ \hline
Ours & \textbf{0.916} & \textbf{0.738}  &\textbf{0.347}\\ \hline
LinkNet \cite{shvets2018automatic} & {0.906} & {0.704} & 0.324\\ \hline
ICNet \cite{zhao2018icnet} & 0.882 & 0.553 & 0.266\\ \hline
UNet \cite{ronneberger2015u} & 0.882 & 0.588 & 0.258\\ \hline
TernausNet \cite{iglovikov2018ternausnet} & 0.835 & 0.587 & 0.263\\ \hline
PSPNet\cite{zhao2017pyramid} & 0.831    & 0.559  &0.232 \\ \hline
\end{tabular}
\end{center}
\end{table}

\subsection{Branch Analysis}
We calculate the speed and accuracy in our auxiliary branch and compare with the main branch. Table \ref{tab:branch_perf} compares the fps and Dice scores of both branches in binary, parts, and instruments wise segmentation. It requires 8x upsample of auxiliary feature maps to measure performance with original ground-truth. As MFF is fusing master branch features with the auxiliary branch, hence it has almost similar performance as a master branch but faster inference time. It can be a trade-off to auxiliary branch instead of the main branch if it needs higher speed.
\begin{table}[!h]
\caption{Performance analysis in different branches of our proposed model}
  \centering
  \renewcommand{\arraystretch}{1.2}
  \begin{tabular}{|c|c|c|c|c|}
    \hline
    \centering
    \multirow{2}{1cm}{{Branch}} & \multirow{2}{.5cm}{{fps}} & \multicolumn{3}{c|}{{Dice}}\\
    \cline{3-5}
    & &{Binary} & {Parts} & {Instruments}\\
    \hline
    Main  &{173.78} &\textbf{0.916} &\textbf{0.738} &\textbf{0.347}\\ \hline
    Auxiliary   & \textbf{227.38} & 0.911 & 0.732 & 0.339\\ \hline
  \end{tabular}
  \label{tab:branch_perf}
\end{table}

\section{Discussion and Conclusion}
In this work, we present a real-time robotic instrument segmentation method based on pixel level semantic segmentation. We propose a multi-resolution feature fusion (MFF) module which can fuse the feature maps with different dimensions and channels. We also adopt spatial pyramid pooling by replacing concatenation operation with summation which ensures the multi-scale contextual features without increasing trainable parameters. We choose an auxiliary branch to extract low-resolution features and provides auxiliary loss to optimize model training. Our adversarial training scheme improves the prediction accuracy by detecting and correcting higher order inconsistencies. 
We compare the real-time performance of our model with the existing state of the art models in terms of segmentation accuracy and inference speed.
However, we trade-off between the speed with accuracy to design an optimized model architecture. Sometimes, we use a decoder or deconvolution layer instead of an up-sampling layer which increases the trainable parameters and model complexity. Hence, our model requires higher trainable parameters and slower comparing to LinkNet and UNet. On the other hand, we replace the concatenation operation with summation and tune the kernel size and number to maintain a light-weight architecture. However, there are still limitations in our model. Case 5 (failure) in Fig. \ref{fig:visual_comparison_binary} appears false positives (light reflection) and false negatives (instruments) in the prediction of all the models. Moreover, in Table \ref{table:all_performance}, it is clear that all the models perform poorly in the segmentation on instrument category. These can be improved by doing further investigation. 

Moreover, Surgical scene understanding in robot-assisted surgery includes the segmentation of tissue as well as instruments. The experimental results suggest that the proposed method is highly optimized for robotic instrument segmentation and can also be applied in tissue segmentation. Thus, our work has incorporated substantial innovations as compared to previous findings and provides a baseline for future work on real-time surgical guidance and robot-assisted surgeries.

\section*{ACKNOWLEDGMENT}
This work is supported by the Singapore Academic Research Fund under Grant {R-397-000-297-114}, and NMRC Bedside \& Bench under grant R-397-000-245-511 awarded to Dr. Hongliang Ren.

\bibliographystyle{IEEEtran}
\bibliography{IEEEabrv,mybib}

\end{document}